\documentclass[a4paper]{article}
\usepackage{INTERSPEECH_v2}
\usepackage[dvipsnames,svgnames,x11names]{xcolor}
\usepackage{cite}
\usepackage{multirow}
\usepackage{graphicx}
\usepackage{tikz}
\usepackage{url}
\usepackage[normalem,normalem,normalbf]{ulem}
\useunder{\uline}{\ul}{}

\definecolor{plum}{rgb}{0.56, 0.27, 0.52}

\usepackage[markup=underlined]{changes}

\title{Semi-supervised acoustic model training for five-lingual code-switched ASR}

\name{Astik Biswas$^1$, Emre Y\i lmaz$^2$, Febe de Wet$^1$, Ewald van der Westhuizen$^1$, Thomas Niesler$^{1}$}
\address{
  $^1$ Department of Electrical and Electronic Engineering, Stellenbosch University, South Africa \\
  $^2$ Department of Electrical and Computer Engineering, National University of Singapore, Singapore}
  
\email{abiswas@sun.ac.za, emre@nus.edu.sg, fdw@sun.ac.za, ewaldvdw@sun.ac.za, trn@sun.ac.za}

\usepackage{todonotes}

\begin{document}
\maketitle
\begin{abstract}
This paper presents recent progress in the acoustic modelling of under-resourced code-switched (CS) speech in multiple South African languages.
We consider two approaches.
The first constructs separate bilingual acoustic models corresponding to language pairs (English-isiZulu, English-isiXhosa, English-Setswana and English-Sesotho).
The second constructs a single unified five-lingual acoustic model representing all the languages  (English, isiZulu, isiXhosa, Setswana and Sesotho).
For these two approaches we consider the effectiveness of semi-supervised training to increase the size of the very sparse acoustic training sets.
Using approximately 11 hours of untranscribed speech, we show that both approaches benefit from semi-supervised training.
The bilingual TDNN-F acoustic models also benefit from the addition of CNN layers (CNN-TDNN-F), while the five-lingual system does not show any significant improvement.
Furthermore, because English is common to all language pairs in our data, it dominates when training a unified language model, leading to improved English ASR performance at the expense of the other languages.
Nevertheless, the five-lingual model offers flexibility because it can process more than two languages simultaneously, and is therefore an attractive option as an automatic transcription system in a semi-supervised training pipeline.
\end{abstract}


\vspace*{-1mm}
\section{Introduction}
\label{sec:intro}
South Africa is a multilingual country with a population of 57 million people and 11 official languages, including English which is used as a \emph{lingua franca} and widespread in the media and entertainment.
Due to this variety of geographically co-located languages, code-switching (CS)---the alternation between languages during discourse---is common.

The development of speech recognition systems able to process code-switching is a topic that has attracted increasing research interest~\cite{modipa2013implications,vu2012first,wu2015code,yilmaz2016code,adel2013recurrent}.
The task becomes more challenging when some of the languages are under-resourced, since small text and acoustic datasets limit the application of state-of-the-art language and acoustic modelling methods.
While English-Mandarin CS speech is the most studied~\cite{li2013improved,zeng2018end,li2013language,vu2012first}, other language pairs such as Frisian-Dutch \cite{yilmaz2016code,yilmaz2017language}, Hindi-English \cite{pandey2018phonetically,emond2018transliteration}, English-Malay \cite{ahmed2012automatic} and French-Arabic \cite{amazouz2019addressing} have also been considered.
Recently, a 14.3-hour corpus of manually segmented and transcribed CS speech has been compiled from South African soap opera episodes \cite{van2018city}. 
This corpus contains four language-balanced South African CS pairs: English-isiZulu (EZ), English-isiXhosa (EX), English-Setswana (ET), and English-Sesotho (ES).

In South Africa, code-switching is most prevalent between English, a highly-resourced language,  and the nine official African Bantu languages,  which  are  all under-resourced. 
We have demonstrated that multilingual training is effective for ASR of bilingual code-switched speech when additional training data consists of closely-related languages~\cite{biswas2018IS}. 
Since isiZulu, isiXhosa, Sesotho and Setswana belong to the same Bantu language family, they were found to complement each other and allow the development of better acoustic models.  
However, 12.2 hours of training data remains insufficient for robust ASR. 
In further work, it was demonstrated that better-resourced but out-of-domain monolingual English and Bantu speech could further improve ASR performance for code-switched speech~\cite{biswas2018improving}. 
Here, approximately an order of magnitude more out-of-domain monolingual speech was available for acoustic model training than in-domain soap opera speech. 
However, the large amount of monolingual, out-of-domain data proved less effective than a modest amount of in-domain code-switched data. 
The out-of-domain data was prompted speech and poorly matched with the in-domain CS speech.
Hence, obtaining more in-domain data remains the ideal solution.

The collection of multilingual code-switched speech from soap opera episodes and its manual segmentation and annotation is extremely intensive in terms of effort and time. 
It has recently been shown that, in the absence of manually-annotated material, automatically-transcribed training material may be useful in under-resourced scenarios using semi-supervised training~\cite{thomas2013deep, yilmaz2018semi, Guo2018}. 
In order to evaluate this for our corpus, we considered the automatic transcription of new and untranscribed soap opera speech using our best  existing code-switched speech recognisers.  

In this study, we investigated semi-supervised training of ASR systems capable of processing all four code-switch pairs in our soap opera corpus.
To the best of our knowledge, this is the first report of semi-supervised training applied to multiple code-switch pairs, and to South African code-switch pairs in particular.
We used our best in-house models, trained on all available in-domain mixed-language and out-of-domain monolingual speech, to automatically transcribe additional segmented but untranscribed code-switched speech.
These segments lack language labels, and were therefore presented to all our code-switched transcription systems as input.
Two such systems were considered.
The first comprises four bilingual code-switched recognisers  while the second is a single unified five-lingual acoustic system. 
The transcriptions produced by both these systems were used to retrain the respective acoustic models. 
For retraining the acoustic models, we used only soap opera speech and no out-of-domain monolingual speech.
We report on the effectiveness of semi-supervised training for both the bilingual and the five-lingual systems.

 \vspace{-6pt}
\section{Multilingual soap opera corpus}
\label{SEC:corpus}
\vspace{-3pt}
A multilingual corpus containing examples of code-switched speech has been compiled from 626 South African soap opera episodes.
The soap opera speech in question is typically fast and often expresses emotion. 
The spontaneous nature of the speech and the high prevalence of code-switching makes it a challenging corpus for ASR. 
The data contains examples of code-switching between South African English and four Bantu languages: isiZulu, isiXhosa, Setswana and Sesotho.
isiZulu and isiXhosa belong to the Nguni language family while Sesotho and Setswana are Sotho languages.

\subsection{Manually transcribed data}
\label{SEC:corpus:transcribed}
\vspace{-3pt}
The soap opera corpus, which is still under development, currently consists of 14.3 hours of speech divided into four language-balanced sets, as introduced in~\cite{van2018city}. 
This corpus was used in some of our previous work~\cite{biswas2018IS, biswas2018improving, yilmaz2018building}.

Approximately 9 hours of manually transcribed monolingual English soap opera speech was available in addition to the language-balanced sets.
This data was initially omitted to avoid a bias toward English.
However, pilot experiments indicated that its inclusion enhanced recognition performance.
The balanced set was therefore pooled with the additional English data for the experiments described here. 
The composition of this larger but unbalanced data set summarised in Table~\ref{tab:duration_unbalanaced_corpora}. 

\begin{table}[h]
\vspace{-8pt}
\caption{Duration in minutes (min) and hours (h) as well as word type and token counts for the unbalanced corpus.}
\label{tab:duration_unbalanaced_corpora}
\centering 
\renewcommand*{\arraystretch}{0.8}
\resizebox{\columnwidth}{!}{
\begin{tabular}{c r r r r r r}
\toprule
{\bf{Language}} & \begin{tabular}[c]{@{}c@{}}\bf{Mono}\\  (m)\end{tabular} & \begin{tabular}[c]{@{}c@{}}\bf{CS}\\ (m)\end{tabular}&
\begin{tabular}[c]{@{}c@{}}\bf{Total}\\ (h)\end{tabular}&
\begin{tabular}[c]{@{}c@{}}\bf{Total}\\ (\%)\end{tabular}&
\begin{tabular}[c]{@{}c@{}}\textbf{Word}\\ \textbf{tokens}\end{tabular} & \begin{tabular}[c]{@{}c@{}}\textbf{Word} \\ \textbf{types}\end{tabular} \\ \midrule
English & 754.96              & 121.81  & 14.61 & 69.26 & 194\,426              & 7\,908              \\
isiZulu & 92.75               & 57.41  & 2.50 & 11.86 & 24\,412              & 6\,789              \\
isiXhosa & 65.13               & 23.83  & 1.48 & 7.03  & 13\,825              & 5\,630              \\ 
Sesotho & 44.65               & 34.04  & 1.31 & 6.22  & 22\,226              & 2\,321              \\ 
Setswana & 36.92               & 34.46  & 1.19 & 5.64  & 21\,409              & 1\,525              \\ \midrule
{\bf{Total}} & 994.43              & 271.54  & 21.10 & 100  & 276\,290              & 24\,170              \\ \bottomrule
\end{tabular}%
}
\end{table}

An overview of the statistics for the development ({\textbf{Dev}}) and test ({\textbf{Test}}) sets for each language pair is given in Table \ref{tab:corpora_stat}.
Besides the total duration, the duration of the monolingual (m) and code-switched (c) segments is included in the table.
A total of 1\,464, 691, 798, and 1\,025 language switches are observed in the EZ, EX, ES, and ET test sets, respectively.

\begin{table}[h]
\scriptsize
	\centering
	\caption{Duration (minutes) of English, isiZulu, isiXhosa, Sesotho, Setswana monolingual (mdur) and code-switched (cdur) utterances in CS development and test sets \cite{van2018city}.}\label{tab:corpora_stat}
	\vspace{-8pt}
    \renewcommand*{\arraystretch}{0.9}
		\begin{tabular*}{0.47\textwidth}{@{\extracolsep{\fill}}c r r r r r @{}}
			\toprule
			\multicolumn{6}{c}{\textbf{English-isiZulu}} \\
			& emdur & zmdur & ecdur & zcdur & \textbf{Total} \\
			\textbf{Dev} & 0.00 & 0.00 & 4.01 & 3.96 & 8.00 \\
			\textbf{Test} & 0.00 & 0.00 & 12.76 & 17.85 & 30.40 \\ \midrule
			\multicolumn{6}{c}{\textbf{English-isiXhosa}} \\
			& emdur & xmdur & ecdur & xcdur & \textbf{Total} \\
			\textbf{Dev} & 2.86 & 6.48 & 2.21 & 2.13 & 13.68 \\
			\textbf{Test} & 0.00 & 0.00 & 5.56 & 8.78 & 14.34 \\ \midrule
			\multicolumn{6}{c}{\textbf{English-Setswana}} \\
			 & emdur & tmdur & ecdur & tcdur & \textbf{Total} \\
			\textbf{Dev} & 0.76 & 4.26 & 4.54 & 4.27 & 13.83 \\
			\textbf{Test} & 0.00 & 0.00 & 8.87 & 8.96 & 17.83 \\ \midrule
			\multicolumn{6}{c}{\textbf{English-Sesotho}} \\
			 & emdur & smdur & ecdur & scdur & \textbf{Total} \\
			\textbf{Dev} & 1.09 & 5.05 & 3.03 & 3.59 & 12.77 \\
			\textbf{Test} & 0.00 & 0.00 & 7.80 & 7.74 & 15.54 \\ \bottomrule
		\end{tabular*}%
	\vspace{-10pt}
\end{table}

\subsection{Manually segmented untranscribed data}
\label{SEC:corpus:untranscribed}
\vspace{-3pt}
In addition to the transcribed data introduced in the previous section, 23~290 segmented but untranscribed soap opera utterances were available for experimentation.
These utterances correspond to 11.05 hours of speech from a total of 127 speakers (69 male and 57 female).
The languages used in these unsubscribed utterances have not been identified.
However, several South African languages not among the five present in the transcribed data are known to occur.

\section{Semi-supervised training}
\begin{figure} [h]
	\centering
	\includegraphics[width=\columnwidth]{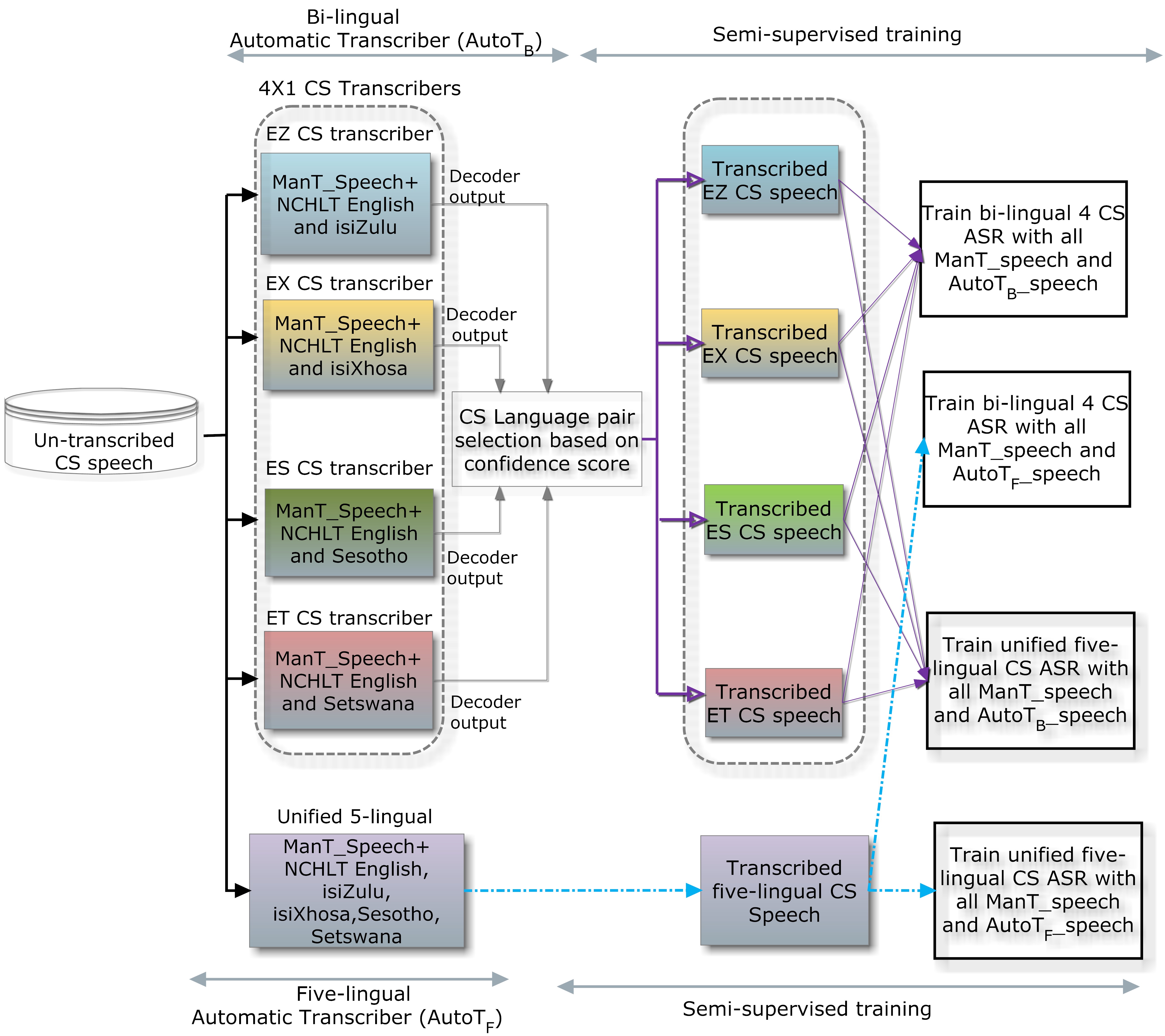}
	\caption{Semi-supervised training framework for the five-lingual  (\protect\tikz[baseline]{\protect\draw[line width=0.3mm,dash dot, cyan, ->] (0,.8ex)--++(1,0)}) and 4$\times$CS (\protect\tikz[baseline]{\protect\draw[line width=0.3mm, plum, -> ] (0,.8ex)--++(1,0)}) transcription systems. (ManT: Manually transcribed data; AutoT: Automatically transcribed data)}
	\label{mdnn}
	\vspace{-8pt}
\end{figure}

A recent study demonstrated that semi-supervised training can improve the performance of Frisian-Dutch code-switched ASR~\cite{yilmaz2018semi}.
We took a similar approach here, using the system configuration illustrated in Figure~\ref{mdnn}. 
The figure shows how semi-supervised training consists of two phases: automatic transcription followed by acoustic model retraining. 
Two different automatic transcription systems were used to transcribe the manually segmented data described in Section~\ref{SEC:corpus:untranscribed}.
Both systems used factorised time-delay neural networks (TDNN-F) as acoustic models, which have recently been demonstrated to be effective in resource-constrained situations~\cite{povey2018}.

 Two different systems were maintained for semi-supervised training and ASR evaluation respectively with the goal of acoustic model development using in-domain data only. 
 This was motivated by the results reported in previous work that similar performance gains can be obtained using a much smaller amount of in-domain training data compared to out-of-domain data \cite{biswas2018improving}. Using less data is also computationally efficient given the reduced amount of automatic transcribed speech data used during the semi-supervised training.

\subsection{Parallel bilingual code-switch transcription}
\vspace{-3pt}
The first transcription system (AutoT$_{B}$) consists of four subsystems, each corresponding to a code-switch language pair~(4$\times$CS). 
Acoustic models were trained on the manually transcribed soap opera data described in Section~\ref{SEC:corpus:transcribed} pooled with monolingual NCHLT speech data for the two languages (on average 52 hours per language) in each pair~\cite{barnard2014nchlt}. 

Because the languages in the untranscribed data were unknown, each utterance was decoded in parallel by each of the bilingual decoders.
The output with the highest confidence score provided both the transcription and a language pair label.
In this way 7\,951 EZ, 3\,796 EX, 11\,415 ES and 128 ET automatically transcribed utterances were obtained. 

\subsection{Unified five-lingual code-switch transcription}
\vspace{-3pt}
The second transcription system (AutoT$_{F}$) was based on a single acoustic model trained on all five languages~\cite{yilmaz2018building}. 
The five-lingual system was trained on the manually transcribed soap opera data introduced in Section~\ref{SEC:corpus:transcribed} pooled with monolingual NCHLT data for all five target languages~\cite{barnard2014nchlt}. 
Since the five-lingual system is not restricted to bilingual output, Bantu-to-Bantu language code-switching was also observed in the transcriptions.
Interestingly, it was also observed that the automatically generated transcriptions sometimes contain more than two languages. 
Although using three different languages within a single utterance is not common, the soap opera corpus does contain such examples.
Of the 23\,290 untranscribed utterances, the five-lingual system classified 3\,390 as isiZulu, 142 as isiXhosa, 657 as Setswana, 1\,069 as Sesotho, 3\,952 as English and 14\,080 as containing code-switching. 

\section{Experiments}
\label{sec:exps}

\subsection{Language modelling}
\label{sec:LM}
The EZ, EX, ES, ET vocabularies respectively contain 11\,292, 8\,805, 4\,233, 4\,957 word types and were closed with respect to the train, development and test sets.
The SRILM toolkit \cite{stolcke2002srilm} was used to train and evaluate all language models.
Table~\ref{perplexity} provides overall development set perplexities and a detailed perplexity breakdown for the test set. 
The code-switch perplexity (CPP) is the perplexity computed only across a language switch, while the monolingual perplexity (MPP) calculation excludes language switches.

\begin{table}[h]
\scriptsize
\caption{Development and test set language model perplexities.  (EB: English to Bantu switch; BE: Bantu to English switch.)}
\label{perplexity}
\centering
\begin{tabular*}{0.47\textwidth}{@{\extracolsep{\fill}} l @{\hspace{4pt}} r @{\hspace{4pt}} r @{\hspace{4pt}} r @{\hspace{4pt}} r @{\hspace{4pt}} r @{\hspace{4pt}} r @{\hspace{4pt}} r @{\hspace{4pt}} r @{} }
\toprule
 & Dev & Test & all CPP & CPP$_{\rm EB}$ & CPP$_{\rm BE}$ & all MPP & MPP$_{\rm E}$ & MPP$_{\rm Z}$ \\
\midrule
\multicolumn{9}{c}{\textbf{Bilingual 3-gram language model}} \\
EZ & 425.82 & 601.69 & 3\,291.95 & 3\,834.99 & 2\,865.41 & 358.08 & 121.15 & 777.76 \\
EX & 352.87 & 788.81 & 4\,914.45 & 6\,549.59 & 3\,785.64 & 459.04 & 96.82 & 1\,355.65 \\ 
ES & 151.47 & 180.47 & 959.01 & 208.61 & 4\,059.13 & 121.24 & 126.87 & 117.84 \\ 
ET & 213.34 & 224.53 & 1070.18 & 317.34 & 3\,798.06 & 160.40 & 142.15 & 176.14 \\
\midrule
\multicolumn{9}{c}{\textbf{Unified five-lingual 3-gram language model}} \\
EZ & 599.93 & 1\,007.15 & 6\,708.18 & 17\,371.00 & 2\,825.16 & 561.80 & 94.45 & 2\,013.00 \\
EX & 669.07 & 1\,881.82 & 15\,083.65 & 50\,208.32 & 5\,058.00 & 1\,015.93 & 87.61 & 5\,590.05 \\
ES & 365.48 & 345.35 & 3\,617.44 & 2\,607.15 & 5\,088.76 & 207.84 & 103.88 & 355.76 \\
ET & 236.96 & 277.48 & 2\,936.63 & 1\,528.38 & 5\,446.35 & 158.15 & 99.76 & 211.25 \\
\bottomrule
\end{tabular*}%
\vspace{-10pt}
\end{table}

\subsubsection{Bilingual language modelling}
\vspace{-2pt}
Each bilingual language model was trained using the respective bilingual training set transcriptions, monolingual English and corresponding Bantu language training texts.
Further details regarding these 3-gram language models are presented in~\cite{biswas2018improving}.

Table~\ref{perplexity} shows that, for each CS pair, the monolingual Bantu perplexity is much higher than the corresponding English value.
This is expected given the much larger text training corpus available for monolingual English than for the Bantu languages (471M vs.\ 8M words respectively). 
The CPP for switching from English to isiZulu/isiXhosa is much higher than when switching from these languages to English.
This can be ascribed to the much larger vocabularies for isiZulu and isiXhosa, which are in turn due to the high degree of agglutination and the use of conjunctive orthography in these languages.

\subsubsection{Five-lingual language modelling}
\vspace{-2pt}
A five-lingual trigram was obtained by interpolating  (1) a trigram trained on all code-switched text, (2) a 4-language trigram trained on all monolingual text from the four African languages, and (3) an English trigram. 
The interpolation weights were optimised on the transcriptions of the development data. 
The five-lingual trigram had overall perplexities of 412 and 617 on the development and test transcriptions respectively. 
The perplexities for each individual code-switch language pair are reported in the Table~\ref{perplexity}. 
It is apparent that the perplexities of the five-lingual trigram are substantially higher than those of the the bilingual language models. 
This is because the five-lingual trigram allows language switches not permitted by the bilingual models. 
Due to their agglutinative orthography, EZ and EX suffer most.
This effect is most pronounced for EX, which has the smallest training set.
On the other hand, English monolingual perplexities are better for the unified model than for the the bilingual language models.
Since English is common to all four CS language pairs, much more in-domain English training data was available. 

\subsection{Acoustic modelling}
\vspace{-2pt}
All ASR experiments were performed using the Kaldi ASR tookit \cite{povey2011kaldi} and the data described in Section~\ref{SEC:corpus}. 
 For  acoustic model retraining, only soap opera speech was used.
A total of 33 hours of speech (including manually  and automatically transcribed data) was used to retrain the acoustic models for both the bilingual and five-lingual South African code-switch speech recognisers. 

For  multilingual  training, the training sets of all the relevant languages were pooled.  
All acoustic models are language dependent, implying that no phone merging was performed between languages.  
First, a context-dependent Gaussian mixture model - hidden Markov model  (GMM-HMM)  acoustic model with 25k Gaussians was trained using 39 dimensional mel-frequency cepstral coefficient (MFCC) features including velocity and acceleration coefficients.  
This GMM-HMM was used to obtain the alignments required for neural network training. 
The same pool of training data was used to derive acoustic features for neural network training.   
Three-fold data augmentation~\cite{ko2015audio} was applied prior to the extraction of MFCCs (40-dimensional, without derivatives), pitch features (3-dimensional) and i-vectors for speaker adaptation (100-dimensional).

Two types of neural network-based acoustic model architectures were evaluated:  (1) the recently proposed TDNN-F models~\cite{povey2018}, which have been shown to be effective in under-resourced scenarios, and (2) TDNN-F with added convolutional layers (CNN-TDNN-F). 
It has recently been shown that the locality, weight sharing and pooling properties of the convolutional layers have potential to improve the performance of ASR \cite{abdel2014convolutional}. 

The TDNN-F models (10 time-delay layers followed by a rank reduction layer) were trained according the standard Kaldi Librispeech recipe (version 5.2.164). 
The CNN-TDNN-F models consisted of 2 CNN layers followed by 10 time-delay layers and a rank reduction layer.
The default hyperparameters of the standard recipes were used and no hyperparameter tuning was performed for neural net training.
For the bilingual experiments, the multilingual acoustic models were adapted to the four different target language pairs. 

\section{Results and discussion}
ASR quality was measured by evaluating the word error rate (WER) on the  EZ, EX, ES and ET development and test sets described in Table~\ref{tab:corpora_stat}.
Note that these test utterances always contain code-switching and are never monolingual. 
The results for different configurations of semi-supervised training  are reported in Table \ref{results1}. 
In this table, AutoT$_B$ indicates that the transcriptions produced by the bilingual transcription system were used to retrain acoustic models, while AutoT$_F$ indicates that transcriptions produced by the five-lingual system were used.  

\vspace*{-3pt}
\subsection{Bilingual semi-supervised experiments}
\vspace*{-1pt}
Table~\ref{results1} shows that, for the TDNN-F acoustic model architecture, one iteration of semi-supervised training using the output of the bilingual system led to an absolute WER reduction of 2.68\% relative to the baseline.
Using a CNN-TDNN-F affords an additional absolute WER reduction of 2.43\%. 

\begin{table}[h]
\vspace{-3pt}
\caption{Mixed WERs (\%) for bilingual and five-lingual ASR. 
}
\vspace{-6pt}
\label{results1}
\resizebox{\columnwidth}{!}{%
\begin{tabular}{@{}lcccccccc@{}}
\hline
\multicolumn{9}{c}{\textbf{Bilingual code-switched ASR}} \\
\multirow{2}{*}{\begin{tabular}[c]{@{}c@{}}CS\\ Pair\end{tabular}} & \multicolumn{2}{c}{\begin{tabular}[c]{@{}c@{}}TDNN-F (Baseline)\\ ManT\end{tabular}} & \multicolumn{2}{c}{\begin{tabular}[c]{@{}c@{}}TDNN-F\\ ManT+AutoT$_B$\end{tabular}} & \multicolumn{2}{c}{\begin{tabular}[c]{@{}c@{}}CNN-TDNN-F\\ ManT+AutoT$_B$\end{tabular}} & \multicolumn{2}{c}{\begin{tabular}[c]{@{}c@{}}CNN-TDNN-F\\ ManT+AutoT$_F$\end{tabular}} \\
\cmidrule(lr){2-3} \cmidrule(lr){4-5} \cmidrule(lr){6-7} \cmidrule(lr){8-9} 
 & Dev & Test & Dev & Test & Dev & Test & Dev & Test \\ \hline
EZ & 41.35 & 47.45 & 39.50 & 44.93 & 38.23 & 44.01 & 36.26 & 43.20 \\
EX & 45.74 & 52.28 & 42.35 & 48.74 & 39.74 & 47.27 & 40.39 & 46.74 \\
ES & 58.59 & 60.16 & 56.34 & 56.17 & 53.96 & 53.58 & 53.80 & 52.86 \\
ET & 54.06 & 51.04 & 51.71 & 50.37 & 48.53 & 45.62 & 46.99 & 45.45 \\
Overall & 49.93 & 52.73 & 47.47 & 50.05 & 45.11 & 47.62 & 44.36 & 47.06 \\ \hline
\multicolumn{9}{c}{\textbf{Unified five-lingual code-switched ASR}} \\
\multirow{2}{*}{\begin{tabular}[c]{@{}c@{}}CS\\ Pair\end{tabular}} & \multicolumn{2}{c}{\begin{tabular}[c]{@{}c@{}}TDNN-F (Baseline)\\ ManT\end{tabular}} & \multicolumn{2}{c}{\begin{tabular}[c]{@{}c@{}}TDNN-F\\ ManT+AutoT$_F$\end{tabular}} & \multicolumn{2}{c}{\begin{tabular}[c]{@{}c@{}}CNN-TDNN-F\\ ManT+AutoT$_F$\end{tabular}} & \multicolumn{2}{c}{\begin{tabular}[c]{@{}c@{}}CNN-TDNN-F\\ ManT+AutoT$_B$\end{tabular}} \\
\cmidrule(lr){2-3} \cmidrule(lr){4-5} \cmidrule(lr){6-7} \cmidrule(lr){8-9} 
 & Dev & Test & Dev & Test & Dev & Test & Dev & Test \\ \hline
EZ & 39.69 & 50.27 & 36.64 & 46.24 & 35.81 & 44.80 & 37.28 & 47.26 \\
EX & 44.52 & 63.64 & 43.57 & 59.86 & 42.17 & 60.13 & 42.26 & 59.22 \\
ES & 54.81 & 50.39 & 53.54 & 48.94 & 53.93 & 48.82 & 51.45 & 48.15 \\
ET & 48.26 & 46.04 & 47.37 & 43.33 & 45.05 & 42.94 & 51.17 & 49.12 \\
Overall & 46.82 & 52.58 & 45.28 & 49.59 & 44.24 & 49.17 & 45.54 & 50.94 \\ \hline
\end{tabular}%
}
\vspace{-7pt}
\end{table}

The acoustic model retrained with the text from the five-lingual system (AutoT$_F$) shows further improvement on the development and test sets. 
This improvement may be due to this systems's ability to transcribe in more than two languages, since the untranscribed soap opera speech does contain such data. 

\vspace*{-3pt}
\subsection{Five-lingual semi-supervised experiments}
\vspace*{-1pt}
The values in Table~~\ref{results1} also show that, when training a five-lingual acoustic model, the CNN-TDNN-F neural network does not achieve a significant WER improvement compared with the TDNN-F baseline. 
This is in contrast to the bilingual system, where \mbox{CNN-TDNN-F} showed a substantial benefit.
It should be remembered, however, that five-lingual recognition is more difficult since it allows more freedom in terms of the permissable language switches.
In this light, the WER achieved by the five-lingual system, which is not far from that achieved by the bilingual system, is promising.
In fact, it is observed that the WER for the ES and ET test sets are better than for the corresponding bilingual systems.
The deteriorated performance for EX and EZ might be attributable to the higher corresponding perplexities shown in Table~\ref{perplexity}, which are in turn due to the agglutinating property of isiZulu and isiXhosa.
Finally, in contrast to the bi-lingual system, training the five-lingual ASR with the AutoT$_B$ transcription does not offer any improvement over the AutoT$_F$ transcription.

\vspace*{-3pt}
\subsection{Language specific WER analysis}
\vspace*{-1mm}
For additional insight, language specific WERs are presented in Table \ref{lswer} for the different semi-supervised training strategies. 
For code-switched ASR, the performance of the recogniser at the code-switch points is an important factor to consider. 
The values in Table~\ref{lswer} reveal that the five-lingual recogniser is more biased towards English (lower WER) than the bilingual CS ASR system, for which the Bantu language WERs are better. 
As pointed out for the language model, the bias of the five-lingual system towards English is due to the much larger proportion of in-domain English training material available when pooling the four CS language pairs.
Furthermore, the accuracy at the code-switch points is better when using the bilingual system.
This is to be expected, since the unified five-lingual system faces a greater ambiguity at code-switch points than the bilingual systems.
 
 \begin{table}[t]
 \Huge
 \caption{Language specific WERs (\%) for English (E), isiZulu (Z), isiXhosa (X), Sesotho (S), Setswana (T) and code-switched bigram accuracy (CBA) (\%) of the different semi-supervised training configurations evaluated on the test set.}
 \label{lswer}
 \renewcommand*{\arraystretch}{1}
 \resizebox{\columnwidth}{!}{%
 \begin{tabular}{cccccccccccccc}
 \toprule
 \multirow{2}{*}{} & \multirow{2}{*}{Acoustic Model} & \multicolumn{3}{c}{EZ} & \multicolumn{3}{c}{EX} & \multicolumn{3}{c}{ES} & \multicolumn{3}{c}{ET} \\
 \cmidrule(lr){3-5} \cmidrule(lr){6-8} \cmidrule(lr){9-11} \cmidrule(lr){12-14}
  &  & E & Z & CBA & E & X & CBA & E & S & CBA & E & T & CBA \\
 \midrule
 \multirow{7}{*}{\begin{tabular}[c]{@{}c@{}}{\rotatebox{90}{\parbox{6ex}{Bilingual CS~ASR}}}\end{tabular}} & \begin{tabular}[c]{@{}c@{}}TDNN-F (baseline)\\ (ManT)\end{tabular} & 41.81 & 51.80 & 28.96 & 45.17 & 57.72 & 22.87 & 52.43 & 66.27 & 21.93 & 41.73 & 57.17 & 32.10 \\
  & \begin{tabular}[c]{@{}c@{}}CNN-TDNN-F\\ (ManT)\end{tabular} & 39.98 & 49.92 & 30.81 & 42.47 & 57.26 & 23.30 & 45.67 & 63.53 & 24.31 & 36.89 & 54.78 & 33.76 \\
  & \begin{tabular}[c]{@{}c@{}}CNN-TDNN-F\\ (ManT+AutoT$_B$)\end{tabular} & 33.02 & 48.77 & 35.59 & 39.16 & 54.26 & 25.18 & 44.27 & 60.84 & 27.07 & 34.64 & 52.87 & 35.32 \\
  & \begin{tabular}[c]{@{}c@{}}CNN-TDNN-F\\ (ManT+AutoT$_F$)\end{tabular} & 36.72 & 48.17 & 32.99 & 38.21 & 53.26 & 26.92 & 43.21 & 60.49 & 27.32 & 35.15 & 52.23 & 34.63 \\
  \midrule
  \multirow{6}{*}{\begin{tabular}[c]{@{}c@{}}{\rotatebox{90}{\parbox{10ex}{5-lingual CS~ASR}}}\end{tabular}} & \begin{tabular}[c]{@{}c@{}}TDNN-F (baseline )\\ (ManT)\end{tabular} & 34.53 & 62.36 & 24.93 & 45.43 & 77.56 & 11.14 & 31.83 & 65.13 & 16.54 & 29.09 & 57.29 & 24.88 \\
  & \begin{tabular}[c]{@{}c@{}}CNN-TDNN-F\\ (ManT)\end{tabular} & 34.40 & 61.58 & 25.82 & 45.60 & 76.50 & 12.01 & 33.95 & 66.37 & 15.41 & 27.87 & 58.03 & 24.68 \\
  & \begin{tabular}[c]{@{}c@{}}CNN-TDNN-F\\ (ManT+AutoT$_B$)\end{tabular} & 33.31 & 57.99 & 25.20 & 39.43 & 74.37 & 11.58 & 31.16 & 61.64 & 22.43 & 27.11 & 63.73 & 17.66 \\
  & \begin{tabular}[c]{@{}c@{}}CNN-TDNN-F\\ (ManT+AutoT$_F$)\end{tabular} & 29.61 & 56.49 & 28.48 & 40.73 & 74.97 & 11.87 & 29.77 & 63.94 & 16.17 & 25.38 & 54.60 & 27.02 \\ \hline
 \end{tabular}
}
 \vspace{-10pt}
 \end{table}

\section{Conclusions}
\vspace*{1mm}
We have evaluated semi-supervised acoustic model training with the aim of improving the performance of an under-resourced code-switched ASR system operating in four South African language pairs. 
Approximately 11 hours of manually segmented but orthographically untranscribed soap opera speech containing code-switching was processed by two automatic transcription systems: one consisting of four separate bilingual code-switched speech recognisers and the other of a single unified five-lingual code-switched recogniser. 
The results indicate that both systems were able to reduce the overall WER substantially. 
However, the unified five-lingual system exhibited a greater bias towards English than the system employing four bilingual recognisers.
We also found that the addition of CNN layers to TDNN-F acoustic models (thus providing a CNN-TDNN-F architecture) provided improved speech recognition for the bilingual systems, but not for the unified five-lingual system.
Despite the added confuseability inherent in decoding five languages, the five-lingual system achieved an error rate that was almost as good as that attained by the bilingual systems.
Future work will focus on incorporating improved automatic segmentation as well as speaker and language diarisation into the pipeline in order to further extend the pool of in-domain training data. 

\section{Acknowledgements}
We would like to thank the Department of Arts \& Culture (DAC) of the South African government for funding this research. We also gratefully acknowledge the support of the NVIDIA corporation for the donation of GPU equipment.
\bibliographystyle{IEEEtran}
\bibliography{astik_refer_new}
\end{document}